# A Survey On Semi-Supervised Learning Techniques

V. Jothi Prakash[1], Dr. L.M. Nithya[2]

[1] *(PG Student, Department of Information Technology, SNS College of Technology, Coimbatore, Tamil Nadu, India)*

[2] *(Professor, Department of Information Technology, SNS College of Technology, Coimbatore, Tamil Nadu, India)*

***ABSTRACT:*** *Semi-supervised learning is a learning standard which deals with the study of how computers and natural systems such as human beings acquire knowledge in the presence of both labeled and unlabeled data. Semi–supervised learning based methods are preferred when compared to the supervised and unsupervised learning because of the improved performance shown by the semi-supervised approaches in the presence of large volumes of data. Labels are very hard to attain while unlabeled data are surplus, therefore semi-supervised learning is a noble indication to shrink human labor and improve accuracy. There has been a large spectrum of ideas on semi-supervised learning. In this paper we bring out some of the key approaches for semi-supervised learning.*

***Keywords -*** *semi-supervised learning, generative mixture models, self-training, graph-based models*

## 1. INTRODUCTION

Machine learning is a wide subfield of artificial intelligence. It involves creating algorithms and methods that allow computers to learn. This ability to learn from experience, analytical observation, and other means, results in a system that can endlessly improve itself and thereby offer increased efficiency. Learning is about generality. There are two types of learning called inductive learning and transductive learning. In inductive learning, the mission is to build a good classifier on the training dataset with the capacity to simplify any unseen data. During the time of training the learner has no knowledge about the test dataset. However, in transductive learning the learner is aware of the test dataset at the time of training and therefore only needs to shape a good classifier that generalizes to this known test dataset.

One common approach is supervised learning. In supervised learning the training dataset comprises of only labeled data. The objective is to learn a function which is able to generalize well on unseen data. The name 'supervised' implies that the learner is provided with the necessary labeled data. Another method is unsupervised learning. In this method only some sample data are offered to the system as observations without any label. Unsupervised learning uses processes that try to find the regular patterns of the data. There is no external tutor for the system to locate the pattern of the model and it is the sole responsibility of the learner to find out the necessary actions. In supervised learning the training dataset is completely labeled and in unsupervised learning, none of the training dataset is labeled.

Semi-supervised learning is the process of finding a better classifier from both labeled and unlabeled data. Semi-supervised learning methodology can deliver high performance of classification by utilizing unlabeled data. The methodology can be used to adapt to a variety of situations by identifying as opposed to specifying a relationship between labeled and unlabeled data from data. It can yield an improvement when unlabeled data can reconstruct the optimal classification boundary. Some popular semi-supervised learning models include self-training, mixture models, graph-based methods, co-training and multiview learning. The success of semi-supervised learning depends completely on some underlying assumptions. So the emphasis is on the assumptions made by each model.

## 2. GENERATIVE MODELS

Generative models are possibly the oldest semi-supervised learning method. It assumes a model, $p(x,y) = p(y) p(x|y)$ where $p(x|y)$ is a recognizable mixture distribution. With large amount of unlabeled data, the mixture components can be recognized; then ideally we only need one labeled example per component to fully determine the mixture distribution. As a generative model, a mixture is obviously inductive, and naturally has a relatively less number of parameters. A sample binary classification problem [9] is shown in Fig. 1.

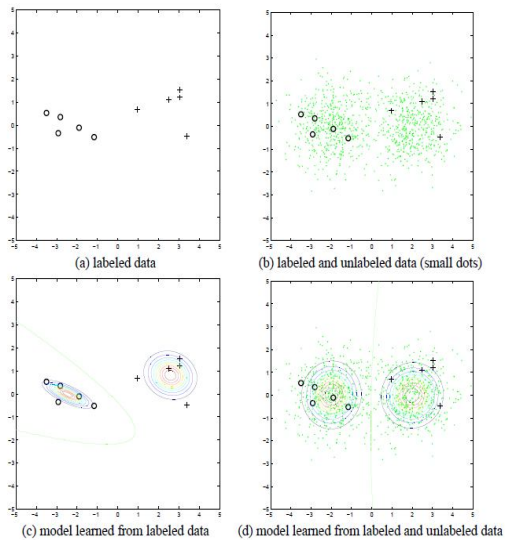

figure 1 binary classification problem

In [1], a new model for bias correction which is similar to the generative model used for training is introduced. The training samples help in estimating the necessary parameters for the bias correction. This generative model is extended by combining a bias correction parameter and discriminative training using the maximum entropy principle. The experiment is evaluated by using three test datasets. The Reuters-21578 dataset containing 135 categories chosen from Reuter's newswire and the ten most recurrent categories is used. The total number of vocabularies is 21,505. The WebKB dataset which contains all the web pages from the universities. There are seven categories in this dataset and out of which only four were chosen which contained 4,199 pages in the categories with 26,389 vocabularies. The 20 newsgroups dataset consisting of 20 unique discussion groups from UseNet with 19,357 vocabularies. The new method was compared with the Naïve Bayes with EM and multinomial logistic regression with minimum entropy regularizer (MLR/MER). The bias correction model outperformed both the methods and at the same time taking care of the overfitting problem.

In [2], Mixture models and graph-based algorithms are combined and spontaneously enforce the smoothness assumption. This approach minimizes the graph problem size and also takes care of the unnoticed test points. Instead of learning all the parameters for the EM method, the harmonic mixture algorithm provides a two-step process. The first step involves training the mixture model using the objective function with standard EM. The second step is fixing the parameters and re-estimating the multinomial to minimize the objective function. One of the main advantages of this method is that the convexity of the objective functions.

The algorithm is tested on the handwritten digits, text categorization, and image analysis tasks. The harmonic mixtures model performs well for all the data and handles the unseen data seamlessly. The mixture model should be identifiable. If the mixture model assumption is correct, unlabeled data is certain to improve accuracy. On the other hand if the assumption by the model is wrong, the unlabeled data may actually degrade the accuracy.

### 3. SELF-TRAINING

Self-training is a commonly used method for semi-supervised learning. In this method a classifier is first trained with the sample of labeled data. Then the classifier is used to classify the unlabeled datasets. Normally the most assured unlabeled data, along with their predicted labels, are appended to the training set. The classifier is re-trained with the new data and the process is repeated. This process of re-training the classifier by itself is called self-teaching or bootstrapping.

In [3], a semi-supervised approach for training object detection systems based on self-training is discussed. The self-training mechanism used is a five-step process. (1) The detector is trained by using a limited set of completely labeled positive samples and a complete set of negative samples. (2) The detector is run over the portion of the dataset with weak labels and the scales and locations are found using the maximum likelihood ratio. (3) The output from the detector is used to label the unlabeled data training samples and a selection score is assigned for each detection. (4) A subset of the newly labeled data is selected using the selection metric. (5) The above steps are repeated until all the data to be trained are added. The method is implemented as a wrapper method over the training process of an existing object detector and the experimental results are provided. This experiential study contributes to the demonstration that a model trained in this way can attain results comparable to a model trained in the traditional way using a much larger dataset of fully labeled data, and that a training dataset selection metric that is defined individually of the detector greatly outperforms a selection metric based on the detection confidence created by the detector. Self-training is a wrapper algorithm, and it is hard to analyse in general.

In [4], two bootstrapping algorithms, Meta-Bootstrapping and Basilisk are discussed. These algorithms are used to exploit mining patterns to learn sets of subjective nouns. A Naive Bayes classifier is trained using the subjective nouns that are exploited, discourse features, and the identified subjectivity clues. A sample of the learned subjective nouns is shown in TABLE 1.

TABLE 1
Example of Learned Subjective Nouns

| Strong Subjective | | Weak Subjective | |
|---|---|---|---|
| tyranny | scum | aberration | plague |
| smokescreen | bully | allusion | risk |
| apologist | devil | apprehensions | drama |
| barbarian | liar | beneficiary | trick |
| belligerence | pariah | resistant | promise |
| condemnation | venom | eyebrows | failures |
| sanctimonious | diatribe | inclination | tolerance |
| exaggeration | mockery | assault | trust |
| repudiation | anguish | blood | spirit |
| antagonism | evil | controversy | slump |
| sympathy | rogue | pressure | rejection |

Meta-Bootstrapping and Basilisk are the algorithms mainly designed for learning semantic words such as apple is a fruit. These two algorithms start with the non-annotated texts and seed words that are semantic in nature. The main working principle of this approach is to identify the subjective words automatically. Meta-Bootstrapping starts with the creation of extraction patterns using the syntactic templates. Then it calculates a score for every pattern based on the seed words and save the best pattern along with all the noun phrases labeled automatically as the semantic category targeted. Meta-Bootstrapping allows only the top-five best nouns phrases to be saved and all the other entries are discarded.

Basilisk uses the extraction patterns to build a sematic dictionary. It also starts with a non-annotated texts and seed words. It involves three steps. (1) Basilisk generates the extraction patterns and scores them automatically and then adds the best patterns to the Pattern Pool. (2) A pattern in the pool extracts all nouns, scores them based on their conformance with the seed words and adds them to the Candidate Word Pool. (3) The top ten nouns are labeled as the target category and are inserted into the dictionary. The bootstrapping algorithms are learned more than a 1000 subjective nouns. Thus self-training paved the way to identify subjective nouns.

## 4. CO-TRAINING

Co-training [5] is a semi-supervised learning technique that needs two views of the data. It assumes that each example is defined using two dissimilar feature sets that provide different, complementary information about the instance. Preferably, the two views are conditionally independent in the sense the two feature sets of each case are conditionally independent and each view is sufficient. The class of an instance can be accurately predicted from each view alone. Co-training starts with learning a separate classifier for each view by using any of labeled samples. The most confident predictions of each classifier on the unlabeled dataset are used to iteratively build further labeled training data.

A co-training style semi-supervised regression algorithm, i.e. COREG, is proposed [6]. This procedure uses two regressors where one regressor labels the unlabeled data for the other regressor. The algorithm used for the regressors in the $k$NN search. The confidence in the labelling of an unlabeled sample is estimated by using the amount of decrease in the mean square error over the labeled region of that sample. In a semi-supervised learning scenario, refining of the regressors take place for all iterations. The kNN method is a lazy algorithm as it does not have a separate phase for training. The algorithm first computes a kNN regressor on the labeled sample and the process stops if there is no more unlabeled sample. The computational cost of the new algorithm in spent mainly on finding the peers of the sample. COREG is compared with the other existing co-training algorithms such as ARTRE, SELF1, and SELF2. In all the cases the performance of COREG is higher than the other algorithms that are compared.

## 5. MULTIVIEW LEARNING

Learning paradigms that employ the agreement among different learners can be defined. The particular assumptions of Co-Training are not required by multiview learning models. Multiview learning models require multiple hypotheses with different inductive biases, e.g., decision trees, Naïve Bayes, SVMs, etc. to be trained from the same labeled dataset, and are necessary to make similar predictions on any given unlabeled instance of data. In [7], semi-supervised learning methods by using two discriminative sequence learning algorithms – the Hidden Markov (HM) perceptron and Support Vector Machines (SVM) a multi-view HM perceptron as well as multi-view 1-norm and 2-norm HM SVMs are developed by consuming the principle of consensus maximization between propositions.

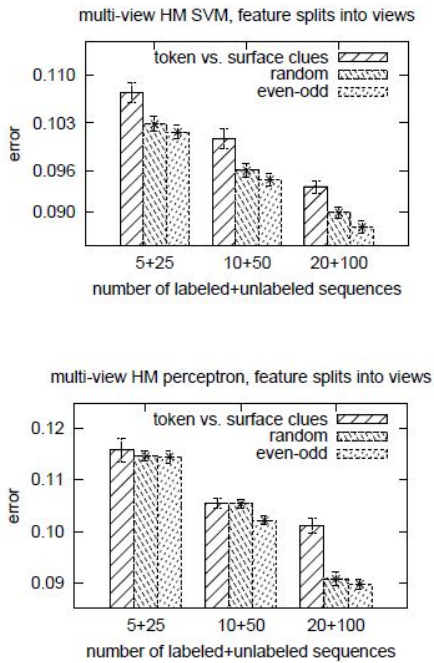

figure 2 error for several splits of features into views

According to the consensus maximization principle, the minimization of the number of errors for the labeled samples has to be minimized by the perceptron algorithm. The label sequence will be predicted by each view for every sample i if it is unlabeled or labeled analogous to the single-view hidden Markov perceptron. The update rules for the labeled samples remain unaltered. If the views do not agree on an unlabeled sample then in order to reduce the disagreement the views have to be updated. Viterbi decoding is used for efficiently computing the joint distribution.

The Hidden Markov SVM is iterated over the samples and successively improves the sample's parameters by using different working set processes for the labeled and unlabeled data. To speed up the computation the difference vectors are removed. If an unlabeled sequence is reached then all the pseudo-sequences of that particular sample are removed as the disagreements in the earlier iterations are resolved. These algorithms perform random feature splits better than splitting the features into a token view and a view of surface clues. The error [7] for several splits of features into views is shown in the Fig. 2.

## 6. GRAPH-BASED MODELS

Graph-based semi-supervised methods define a graph where the nodes are represented as the labeled and unlabeled samples in the dataset, and the edges portray the similarity between the samples. These methods usually assume label smoothness over the graph. Graph based methods require no parameter. These methods are discriminative, and also transductive in nature.

In [8], a general framework for semi-supervised learning on a directed graph is proposed. The structure of the graph along with the direction of the edges is considered. The algorithm takes the input as the directed graph and the label set. The unlabeled instances are classified using the steps. (1) A random walk over the graph with a transition probability matrix is defined such that it has a unique stationary distribution like the teleporting random walk. (2) Calculate the matrix by using the diagonal matrix with its stationary distribution. (3) A function is computed using the labeled vertices to classify the unlabeled vertices. In the absence of labeled instances, this method can be used as a spectral clustering method for the directed graphs. This simplifies the spectral clustering approach for the undirected graphs.

## 7. CONCLUSION

In this survey only a few of the several semi-supervised learning approaches are discussed. As mentioned earlier labeled data is expensive and hard to obtain. On the other hand unlabeled data is comparatively easy to gather. Semi-supervised learning can be used to classify the unlabeled data and also it can be used to develop better classifiers. Semi-supervised learning needs less human labour and gives a better performance than the unsupervised and supervised counterparts. Because of this advantage semi-supervised learning is of great interest in theory as well as in practice.